\pdfoutput=1

\documentclass[final]{cvpr}

\usepackage{times}
\usepackage{epsfig}
\usepackage{graphicx}
\usepackage{amsmath}
\usepackage{amssymb}
\usepackage{verbatim}

\usepackage{array}
\usepackage{xcolor}

\usepackage[pagebackref=true,breaklinks=true,colorlinks,bookmarks=false]{hyperref}

\def\cvprPaperID{****} 
\def\confYear{CVPR 2021}

\begin{document}

\title{Vehicle Re-identification Method Based on Vehicle Attribute \\
and Mutual Exclusion Between Cameras}

\author{Junru Chen\textsuperscript{1} , Shiqing Geng\textsuperscript{2}, Yongluan Yan\textsuperscript{3}, Danyang Huang\textsuperscript{4}, Hao Liu\textsuperscript{4}, Yadong Li\textsuperscript{4}\\
Xidian University\textsuperscript{1} \\
Harbin Engineering University\textsuperscript{2} \\
Huazhong University of Science and Technology\textsuperscript{3} \\
Beihang University\textsuperscript{4}
}

\maketitle

\begin{abstract}
Vehicle Re-identification aims to identify a specific vehicle across time and camera view. 
With the rapid growth of intelligent transportation systems and smart cities, vehicle
Re-identification technology gets more and more attention. However, due to the difference of
shooting angle and the high similarity of vehicles belonging to the same brand, vehicle re-identification 
becomes a great challenge for existing method \cite{8915694}. In this paper, we propose a vehicle attribute-guided 
method to re-rank vehicle Re-ID result. The attributes used include vehicle orientation and vehicle brand . We also focus on the 
camera information and introduce camera mutual exclusion theory to further fine-tune the search results. 
In terms of feature extraction, we combine the data augmentations of multi-resolutions with the large model 
ensemble to get a more robust vehicle features. Our method achieves mAP of 63.73\% and rank-1 accuracy 
76.61\% in the CVPR 2021 AI City Challenge.
\end{abstract}

\section{Introduction}

Given a query vehicle image, vehicle re-identification (Re-ID) aims to find the matched
vehicle from a gallery of vehicle images. Most of existing methods use Convolutional
Neural Network (CNN) \cite{simonyan2015deep, xie2017aggregated, zhu2021aaformer} to extract features of vehicle images and compute the distance between
features to get a rank list of similarity between the query image and gallery images.
Though many existing Re-ID methods \cite{zou2020joint, zhu2020identityguided, li2018diversity, khorramshahi2020devil} achieve great results, there are still several
challenges in the Re-ID task of the city surveillance scene.

One of the challenges is
that the appearance of two images captured by different cameras could be quite different
even though they belong to the same vehicle. The big difference of images belonging to 
the same vehicle comes from the difference of illumination and vehicle orientation. We 
mainly focus on vehicle orientation in this paper.
We introduce a vehicle orientation model to predict the vehicle orientation (0-360 degrees). 
For the images of vehicles with large angle difference, we subtract a value from their 
distance to make them closer in embedding space, so as to improve the accuracy of model prediction. 
Since appearance features of the vehicles facing forward and those facing backward are 
similar, we propose to a folding operation of the angle, which regards the forward and backward 
as the same orientation.

Another challenge is that the appearance of some vehicles with the same brand 
and same type could be quite close. Thus we introduce a coarse vehicle classification 
model for predicting the type of vehicles and a vehicle main brand prediction model 
to assist the vehicle Re-ID task.

\begin{figure}[t]
   \begin{center}
   \fbox{\includegraphics[width=0.45\textwidth]{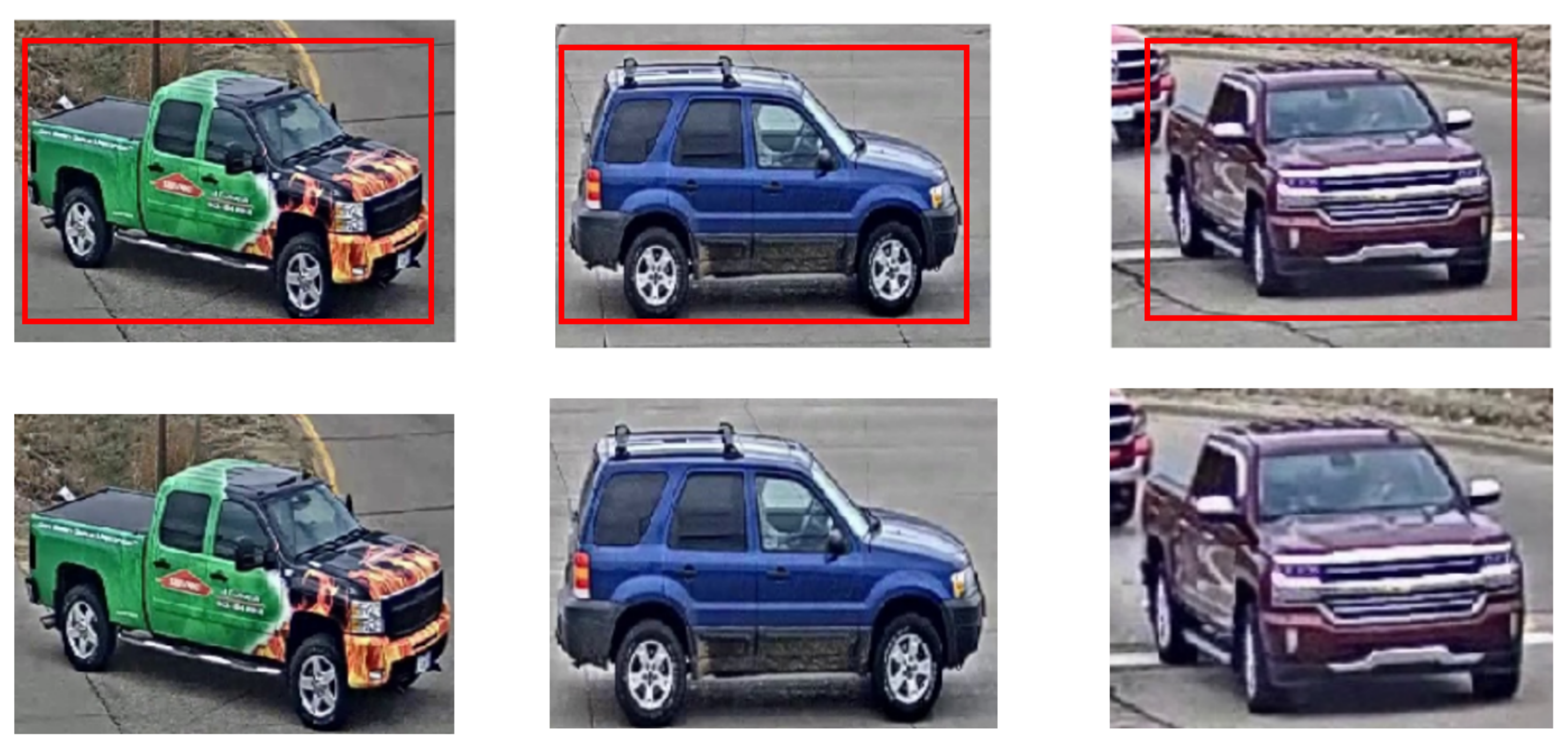}}
   
   \end{center}
      \caption{Visualization of the real-world training data. The first row shows the original images, 
      red bounding boxes indicate the re-detect results. The second row shows the images croped by re-detect results}
   \label{fig: dataset}
\end{figure}

\begin{figure*}[t]
   \begin{center}
   \fbox{\includegraphics[width=0.75\textwidth]{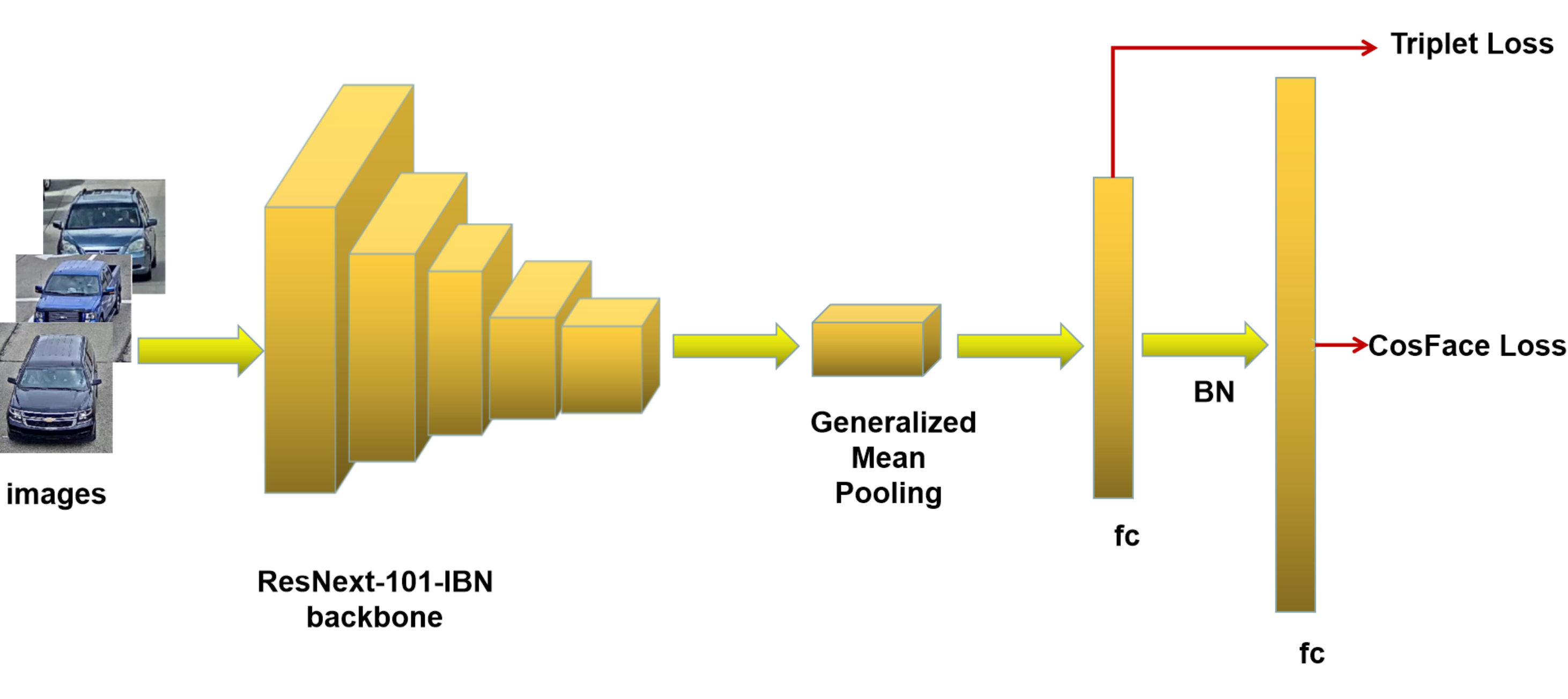}}
   
   \end{center}
      \caption{General architecture of our ResNeXt-101-IBN feature extractor, we normalize 
      the features which output by BN layer as the final retrieval feature.}
   \label{fig: backbone}
\end{figure*}

In addition to using traditional re-rank methods \cite{zhong2017reranking, zhang2020understanding}, we also introduce a re-rank method based on 
the mutual exclusion of camera location. Through observation, we found that the same vehicle 
does not appear in the same camera twice. Therefore, when finding matched gallery image for a query image according to the rank list of similarity, if a image belonging to a certain camera has been retrieved, we move other images captured by this camera to the back of rank list. In other words, we reduce the similarity between the query image and other images belonging to the retrieved camera.

Besides, another challenge in AI City Challenge is that there are many blurred images in 
the test dataset, which makes the model difficult to distinguish images. To address 
above issue, we combine four models with same architecture and different input image size. 
In this way, the whole network can adapt to images of different resolutions, which is equivalent 
to both clear and blurred images. And the real-word training dataset also has the problem of inaccurate 
detection bounding box, so we re-detect those images and crop them for reducing the influence of 
background information, as shown in Figure \ref{fig: dataset}.

\begin{figure*}[t]
   \begin{center}
   \fbox{\includegraphics[width=0.75\textwidth]{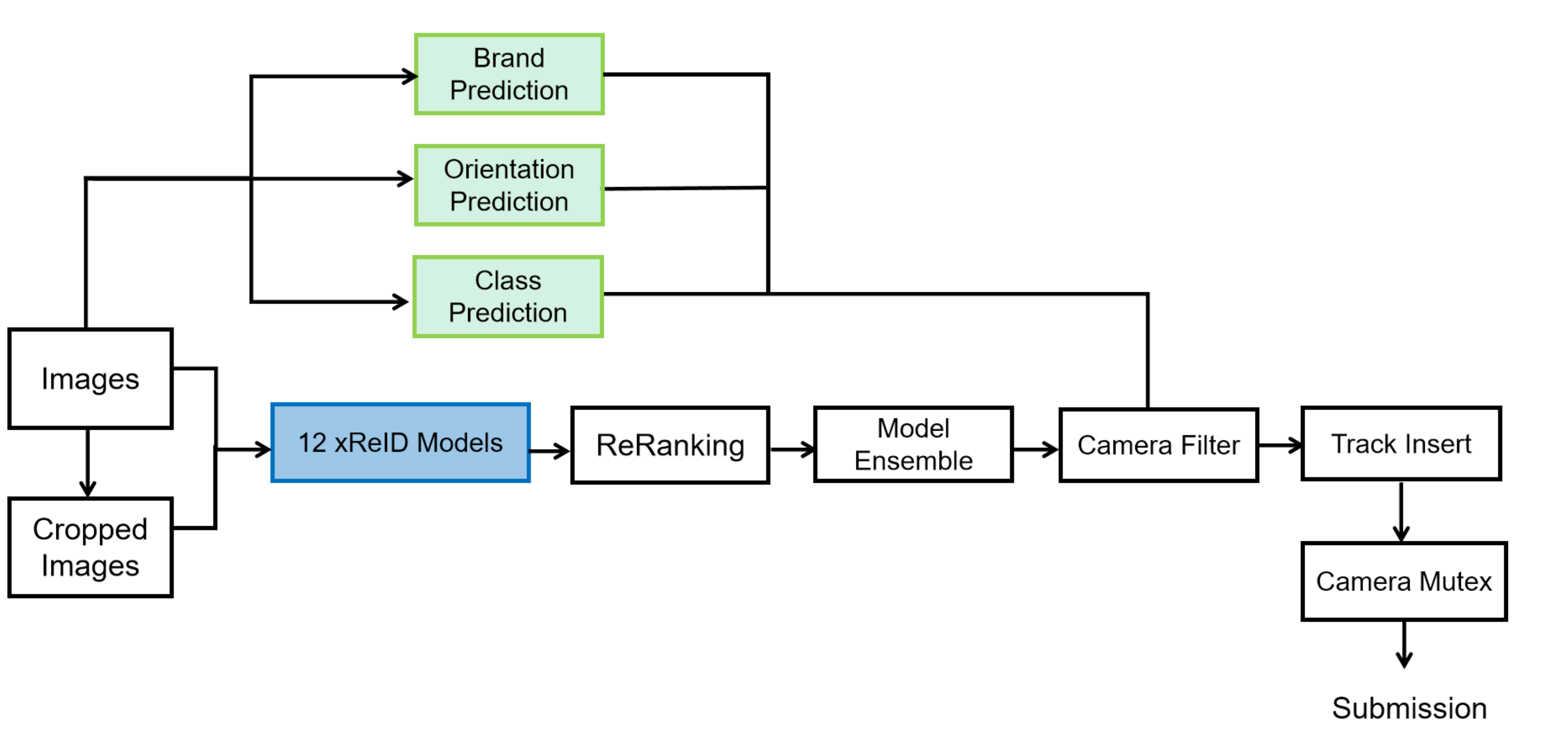}}
   
   \end{center}
      \caption{Testing pipeline. First, the original picture will go through a detection model to obtain the 
      image after removing the redundant background. Send both types of images into 12 feature extractors to get 24 similarity 
      matrices. Re-ranking and adding the matrices to get the ensembled results, filted by the camera id, vehicle orientation, vehicle brand and 
      vehicle type to adjust the similarity matrix. Merge the results of the vehicles which from the same track, those images is inserted 
      behind the top-ranked one. Then use the camera mutual exclusion strategy to get the final search result.
      }
   \label{fig: pipeline}
\end{figure*}

In this paper, our proposed methods mainly focus on how to use large models to 
extract more robust features, the general architecture is shown in Figure \ref{fig: backbone}, and use vehicle orientation and vehicle brand information 
to fine-tune the original ranking result. In summary, our contributions are:
\begin{itemize}
\itemsep=-3pt
    \item\small{We propose a method utilizing vehicle orientation and brand information for vehicle ReID.}
    \item\small{We train the model with multi-scale images and reduce the background area of images using vehicle detection to improve the robustness of model since the quality of images is largely different.}
    \item\small{We propose a camera mutual exclusion theory which combines the query/gallery camera id to fine-tune final retrieval result.}
    \item\small{Our vehicle ReID method achieves mAP of 63.73\% and rank-1 accuracy 76.61\% in the CVPR 2021 AI City Challenge.}
\end{itemize}

\section{Related Work}
With the rapid development of convolutional neural networks(CNNs) \cite{he2015deep, hu2019squeezeandexcitation, pan2020once, zhang2020resnest}, vehicle 
re-identification has made great achievements in recent years. \cite{liu2016deep} proposed
DRDL to map vehicle images into an Euclidean space and used the L2 distance 
between two images for similarity estimation. \cite{liu2018ram} combined a Region
branch which encourages the deep model to learn more local information and thus get more discriminative features. 

Similar to person re-identification \cite{zhang2017alignedreid, sun2018beyond} and 
face recognition \cite{wang2018cosface, schroff2015facenet}, vehicle re-identification 
also aims to learn a feature space which keeps the samples from the same class close 
to each other and those from different classes far away via some effective deep metric 
loss. The commonly used loss functions include Cosface \cite{wang2018cosface}, Circle Loss \cite{sun2020circle}, 
Triplet Loss \cite{schroff2015facenet}. However, there are two main challenges of 
vehicle ReID: large intra-class difference and small inter-class similarity due to 
the viewpoints, cameras, illumination and other factors. Many recent vehicle ReID works 
have focused on learning a robust feature with vehicle attributes, vehicle orientation, 
vehicle keypoints and etc. \cite{chu2019vehicle} considered the viewpoint variation problem 
in vehicle ReID, and proposed a VANet that learns two metrics of similar viewpoints and 
different viewpoints in two feature spaces. \cite{he2019part} introduced a part-regularized 
ReID approach that uses a part-localization network to detect ROIs(Region of interest) and 
then projects the ROIs into the global feature map to learn part features and distinguish 
the subtle discrepancy. In \cite{wang2017orientation}, it combines more local region features 
by presenting 20 fixed key points for vehicle. \cite{wang2020attribute} proposed AGNet with attribute-guided attention module 
which makes full use of vehicle attributes to distinguish different vehicles.

Recently, many transformer-based methods \cite{dosovitskiy2020image, dascoli2021convit, li2021localvit} have been verified in the visual task.
In Re-ID task, there is also a purely transformer based method \cite{he2021transreid, zhu2021aaformer}, and it achieves even better performance than CNN-based methods 
on some human and vehicle re-identification benchmarks.

\section{Method}
In order to get better retrieval results, we have adopted two types of processing methods, 
one type of methods are some general approaches for improving the results of vehicle re-identification, the other type of methods are relatively tricky and aim to improve the performance on AI City Challenge. 
The general methods include re-ranking, model ensemble, and combining additional vehicle information such as 
vehicle type, brand, orientation to fine-tune results. The tricky methods include filtering the 
images of the same camera in gallery, merging the results of the vehicles that from the same 
track, re-ranking based on mutual exclusion between cameras, etc. The flow chart of the entire testing pipeline is shown in 
Figure \ref{fig: pipeline}.

\subsection{Data Preprocessing}
We find that the background area of some images in the \emph{CityFlow} \cite{tang2019cityflow} dataset is too 
large. Therefore, we pass all the training set and test set images through the detection
model to obtain new images for training and testing, experimental results shows that the 
using new detected images can achieve better performance than using the original images.

\begin{figure*}[h]
   \begin{center}
   \fbox{\includegraphics[width=0.9\textwidth]{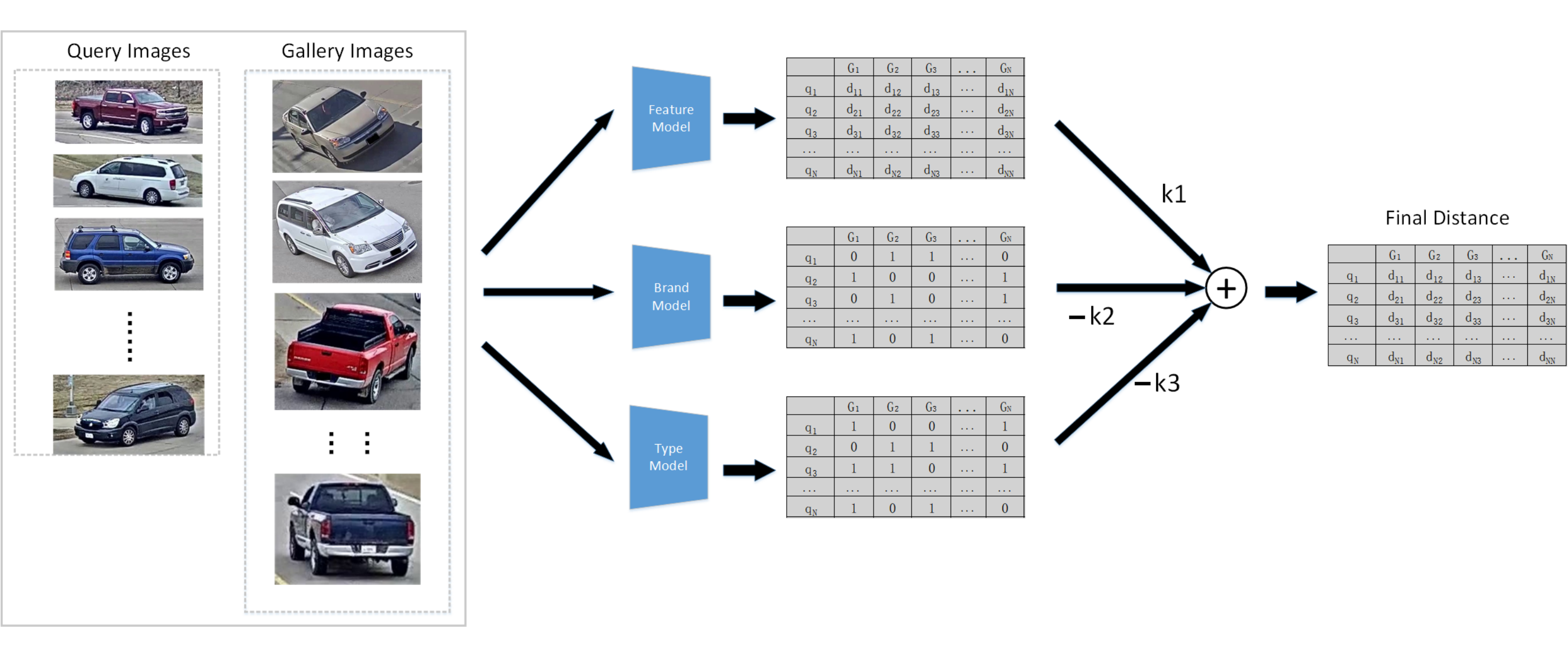}}
   
   \end{center}
      \caption{Vehicle Re-ID integrates vehicle attribute(brand/type) information. 
      When the qury and gallery image pairs have the same vehicle attributes(brand/type), the value of the 
      corresponding position in the matrix is set to 1 (otherwise the value is 0),
      The query gallery image pair has the same attributes, and their distance will be reduced after fusion.}
   \label{fig: vehicle reid with brand attribute}
\end{figure*}

\subsection{Loss Function}
We adopt triplet loss and Cosface loss jointly to optimize the model. The purpose of Cosface
loss is to convert the absolute distance of vectors in the euclidean space into the 
relative distance of the angle between the two vectors in the cosine space. The Cosface loss 
is computed as:

\begin{equation}
   \begin{aligned}
& L_{id}= \\
& \frac{1}{N} \sum_{i}-\log \frac{e^{s\left(\cos \left(\theta_{y_{i}, i}\right)-m\right)}}{e^{s\left(\cos \left(\theta_{y_{i}, i}\right)-m\right)}+\sum_{j \neq y_{i}} e^{s \cos \left(\theta_{j, i}\right)}}
   \end{aligned}
\end{equation},
where $m$ refers to margin of the cosine distance, $s$ means the scale of vectors. In 
experiment, $m$ is set to 0.35 and $s$ is set to 30.

The purpose of triplet loss is to reduce the distance between the features of same 
vehicles and expand the distance between the features of different vehicles. Triplet loss 
is computed as:

\begin{equation}
L_{\text{triplet}} = \left[d_{p} - d_{n} + \alpha\right]_{+}
\end{equation},
where $d_p$ and $d_n$ are the distance of positive pairs and negative pairs in the 
feature space respectively, $\alpha$ is the margin of triplet loss, and $[z]_+$ equals 
to $max(z, 0)$. In our experiments, we set $\alpha$ to 0.5.

Finally, the total loss is computed as:

\begin{equation}
L_{all} = L_{id} + L_{triplet}
\end{equation}

\subsection{Pooling Method}
We found that generalized-mean(GeM) 
pooling \cite{berman2019multigrain} performs better in this task, so we replace the global 
average pooling with GeM at the end of the backbone, GeM is defined as:

\begin{equation}
   \begin{aligned}
& f^{(g)} = \left[f_1^{(g)} \ldots f_k^{(g)} \ldots f_K^{(g)}\right]^T , \\
& f_k^{(g)} = \left(\frac{1}{|X_k|} \sum_{x \in X_k} x^{p_k} \right)^{\frac{1}{p_k}}
   \end{aligned}
\end{equation},
we also try to concat the global average pooling and global max pooling, and fast global average pooling \cite{he2020fastreid}, 
but both of them don't work.

\subsection{Multi-Scale Model Training}
We train four models with same architecture, each of which has a different input image size 
to make the whole network learn multi-scale information. The input image size of the four models 
are $320\times320$, $384\times384$, $416\times416$. For an image from original dataset, we first shrink it to a random size, and then resize the shrinked image to four kinds of sizes mentioned above. In this way, we get four scaled images and we can use them to train the four models respectively. Since the whole network is trained with both blur images and clear images, it can distinguish the blurred image better.

\section{Post-Processing}

\subsection{Vehicle Class/Brand Filter}

Generally the similarity of vehicle images is greatly affected by the camera's
shooting position. For some vehicles of different types or brands, the appearance 
similarity of pictures may still be extremely high. Therefore, we introduce a coarse 
vehicle classification model and a vehicle main brand prediction model to assist the 
vehicle Re-ID task. We use the synthetic data provided by the competition to train the 
vehicle coarse classification model. At the same time, we mark main brands of some vehicles 
on the training set of real vehicles and use them to train the vehicle brand model. 
During the test, we predict the vehicle type and 
brand of the query and gallery images. We increase the similarity of image pairs which 
have the same type of vehicle brand or vehicle class to improve the accuracy of vehicle Re-ID, 
this process is shown in 
Figures~\ref{fig: vehicle reid with brand attribute}.

\begin{table}[h]
   \begin{center}
   \begin{tabular}{|c|c|c|}
   \hline
   {method} & {mAP} & {top-1} \\
   \hline
         ResNeXt-101-IBN         & 0.6147 & 0.7024 \\
         + Re-ranking          & 0.6584 & 0.7551 \\
         + Attribute        & 0.6847 & 0.7988\\
         + Camera Multex       & 0.6943 & 0.8051\\
         + Ensemble  & \textbf{0.7263} & \textbf{0.8344} \\
   \hline
   \end{tabular}
   \end{center}
   \caption{Ablation Study of Post-processing Part in validation data}
   \label{tabel: post-processing}
\end{table}

\begin{table*}[h]
   \begin{center}
   \begin{tabular}{|c | c c c c c c c |}
   \hline
   {ResNeXt-101-IBN} &  \checkmark & \checkmark  & \checkmark  & \checkmark &\checkmark& \checkmark& \checkmark \\
   {Re-ranking}      &             & \checkmark  &             &            &\checkmark& \checkmark& \checkmark \\
   {Attribute}       &             &             & \checkmark  &            &\checkmark&           & \checkmark \\
   {Camera Multex}   &             &             &             & \checkmark &          &\checkmark & \checkmark \\
   {Ensemble}        &             &             &             &            &          &           & \checkmark \\
   \hline

   {mAP}             & 0.6147      & 0.6584     & 0.6412       & 0.6782    &   0.6847  &   0.6804  & \textbf{0.7263} \\
   {rank-1}          & 0.7024      & 0.7551     & 0.7358       & 0.7664    &   0.7988  &  0.7894   & \textbf{0.8344} \\

   \hline
   \end{tabular}
   \end{center}
   \caption{The performance of the combination of different post-processing strategies in validation data, the backbone is ResNeXt-101-IBN.}
   \label{tabel: ablation}
\end{table*}

\subsection{Fine-Tune With Vehicle Orientation}
It is more difficult to search between the vehicles that face front or back and the vehicles that face side
than to search between the vehicles that face front and the vehicles that face back. Therefore, in order to make the 
difficulty of vehicle re-identification consistent in all situations, we artificially reduce the difficulty of mutual searching between the front/back facing and the side facing vehicles.

\begin{figure}[h]
   \begin{center}
   \fbox{\includegraphics[width=0.4\textwidth]{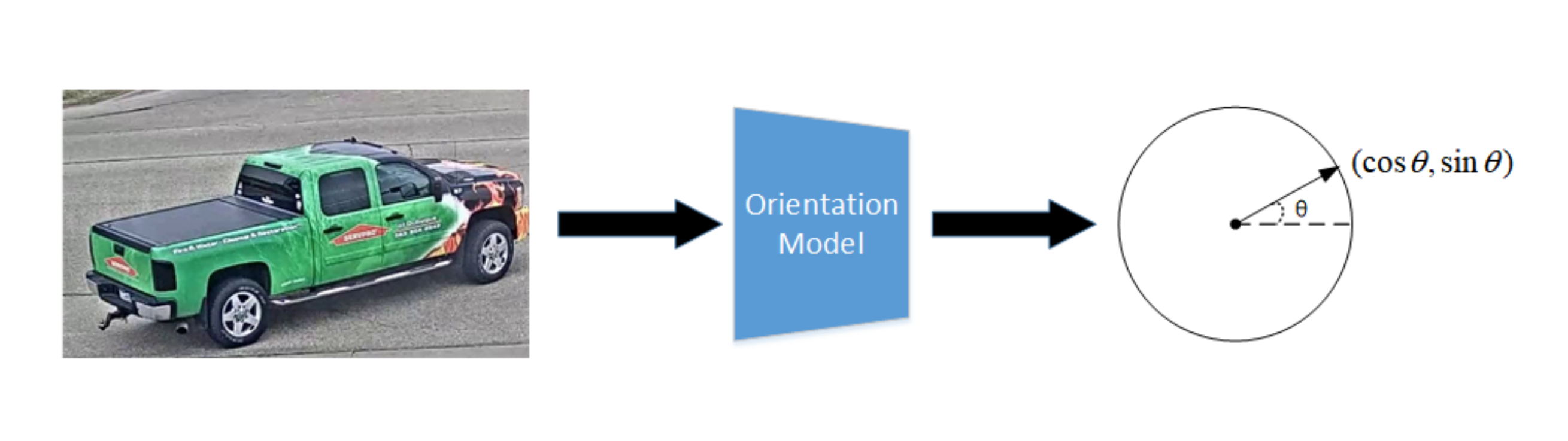}}
   
   \end{center}
      \caption{Model predicts vehicle orientation}
   \label{fig: orientation model output}
\end{figure}

We use the provided synthetic data to train the vehicle orientation prediction model. We divide 0~360 degrees 
into 36 equal parts to get a label of orientation, and train a 36-category classification model. We use the trained model to predict the direction of the vehicle in the test set, and use a unit vector whose direction is consistent with the vehicle orientation to represent the feature of orientation, this process is shown in 
Figures \ref{fig: orientation model output}.

\begin{figure}[h]
   \begin{center}
   \fbox{\includegraphics[width=0.2\textwidth]{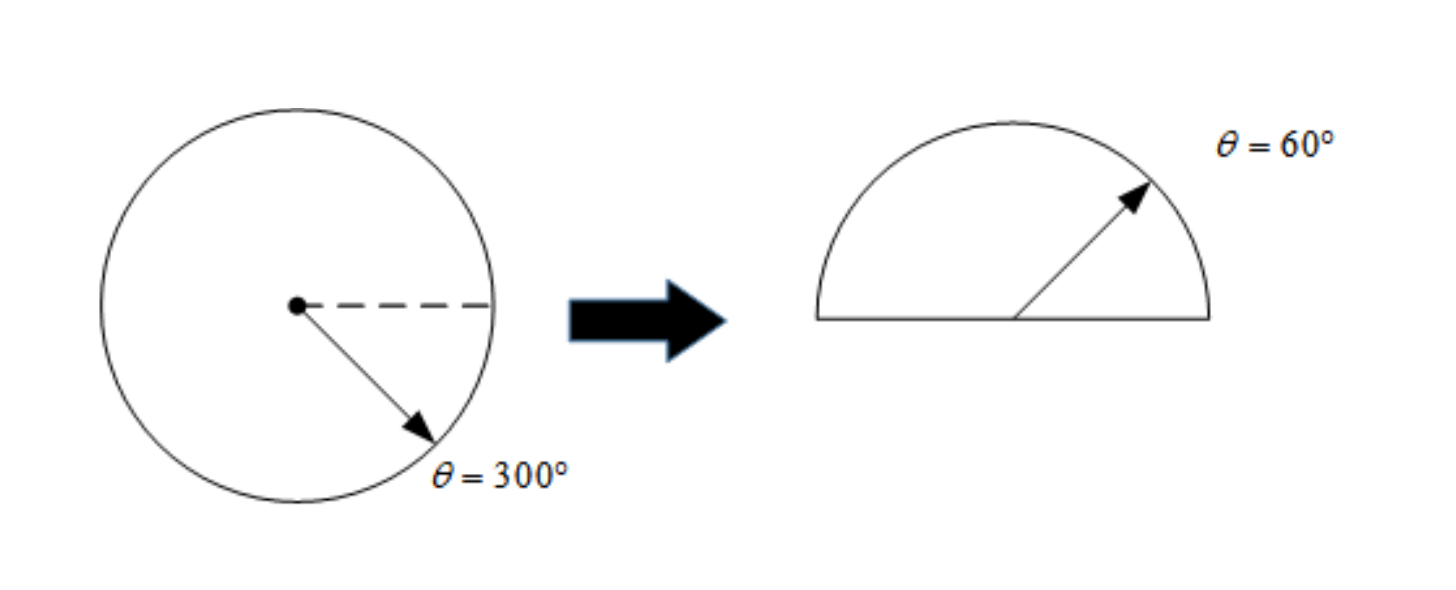}}
   
   \end{center}
      \caption{Fold the vehicle orientation result horizontally}
   \label{fig: angle flip}
\end{figure}

Since it is easy to search between front facing vehicles and back facing vehicles, we fold the vehicle orientations result horizontally, as shown in the Figures \ref{fig: angle flip}, so that the feature similarity between front/back facing vehicle and side facing vehicle will be less than it between front facing and back facing vehicles. In practice, we subtract the similarity matrix of orientation 
from the similarity matrix of re-id with a certain weight $\lambda$.

\subsection{Camera Mutual Exclusion}
In order to make full use of the extra information in the test set, we utilize the camera information and propose the camera mutual exclusion strategy. It can be summarized into two aspects. From the perspective of query-to-gallery, when an image A retrieves the first image from camera $\tau$, following the retrieval order, the remaining 
images which from camera $\tau$ will be artificially moved out retrieval results. From the perspective of gallery-to-query, we find that the vehicles with the same identify in the query all come from different cameras. So when one gallery image $I_{g}$ is the nearest neighbor of multiple query images that are from the same camera, we take only the closest query image as the  $I_{g}$ 's positive sample, and the other query images will be moved out retrieval results.

\begin{table}[h]
   \begin{center}
   \begin{tabular}{|c|c|c|}
   \hline
   {Rank} & {Team ID} & {Score} \\
   \hline
       1     & 47    & 0.7445 \\
       2     & 9     & 0.7151 \\
       3     & 7     & 0.6650 \\
       4     & 35    & 0.6555 \\
       5     & {\textbf{125(Ours)}} & \textbf{0.6373} \\
       6     & 44    & 0.6364 \\
       7     & 122   & 0.6252 \\
       8     & 71    & 0.6216 \\
       9     & 61    & 0.6134 \\
       10    & 27    & 0.6083 \\
   \hline
   \end{tabular}
   \end{center}
   \caption{Performance Evaluation of Track2 in final test data.}
   \label{finalresult}
\end{table}

\begin{figure*}[h]
   \begin{center}
   \fbox{\includegraphics[width=0.9\textwidth]{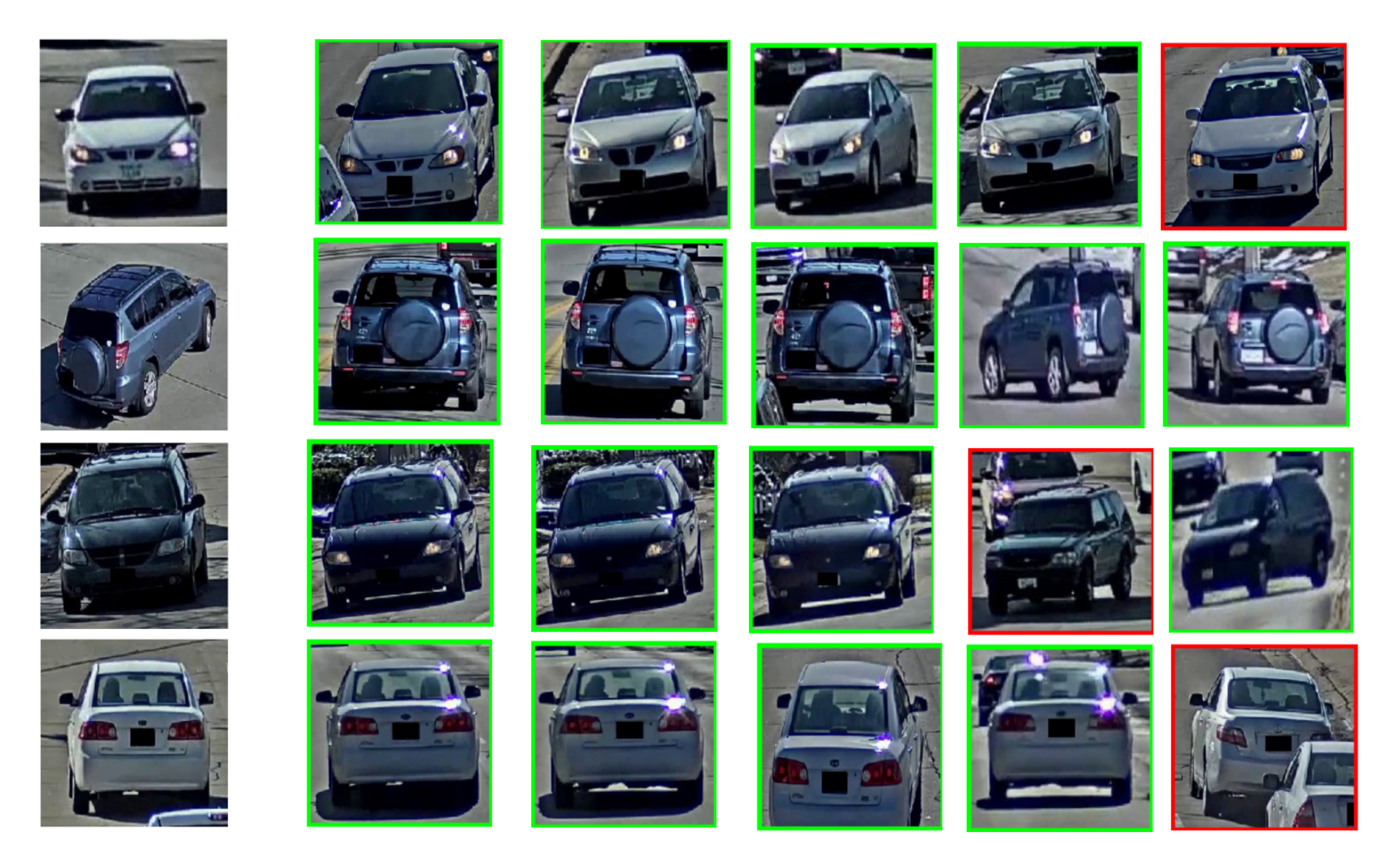}}
   
   \end{center}
      \caption{Visualization of the final retrieval results. The first column shows 
      the query images captured by different cameras, and each row shows the top 5 gallery images 
      retrieved from left to right according to the similarity score. The images in green boxes are 
      true positives, while the images in red boxes are false positives.}
   \label{fig: example}
\end{figure*}

\section{Experiments}
\subsection{Implementation Details}
We adopt IBN-ResNet-101, IBN-ResNeXt-101, IBN-SE-ResNet101 as the backbone networks. 
Given the features extracted by the backbone network, most of Re-ID networks use global 
average pooling(GAP) to obtain a feature vector. 

Following a strong Re-ID baseline \cite{luo2019bag}, we add an BNNeck layer after the backbone. 
In the inference stage, we choose the feature after the BNNeck layer for person id 
prediction. In the training stage, the feature before BNNeck layer is used to 
compute triplet loss and the feature after BNNeck layers is used to compute CosFace 
loss, this process is shown in Figure~\ref{fig: backbone}. 
We train our model with SGD optimizer, setting the momentum to 0.9. The initial learning rate is set to 0.002 and we adopt the cosine strategy to decay the learning rate.

\subsection{Performance Evaluation of Challenge Contest}
We report our challenge contest performance of the track2: City-Scale Multi-Camera 
Vehicle Re-identification. In track2, we won 5th place among all the teams with the 
mAP of 0.6373, as shown in the Table \ref{finalresult}. The actual retrieval results 
of some queries in the contest are shown in Figure \ref{fig: example}. The ablation study of post-processing is shown in 
Tabel \ref{tabel: post-processing} and Table \ref{tabel: ablation}. The ablation study shows that the vehicle re-identification 
combined with attribute (brand/type/orientation), re-ranking and models ensemble have a 
great improvement in model performance. In contrast, the camera mutual exclusion strategy  
improves the model performance slightly.

\section{Conclusion}
In this paper, we propose several effective methods to achieve remarkable 
results in the CVPR 2021 AI City Challenge. We utilize the orientation and 
brand information of vehicles to improve the performance of vehicle ReID. 
We introduce a multi-scale model training method to make the model adapt 
to both blurred and clear images. Since the background areas of some images 
are quite large, we introduce vehicle detection to cut the vehicle part from 
original images in the dataset. We utilize mutual exclusion between cameras to 
optimize the final retrieval results. Finally, our proposed system rank number 5 
(team ID 125) among all the teams with the mAP of 0.6373 for City-Scale Multi-Camera 
Vehicle Re-identification.

{\small
\bibliographystyle{ieee_fullname}
\bibliography{egbib}
}

\end{document}